\documentclass[letterpaper, 10 pt, conference]{ieeeconf}  % Comment this line out if you need a4paper

\IEEEoverridecommandlockouts                              % This command is only needed if 
\usepackage{graphicx}
\usepackage{tabularx}
\usepackage{cite}
\usepackage{algorithmic}
\usepackage{graphicx}
\usepackage{subfigure}
\usepackage{textcomp}
\usepackage{xcolor}
\usepackage{amssymb}
\usepackage{bbding}
\usepackage{multirow}
\usepackage{array}

\title{\LARGE \bf
A Novel and Efficient Tumor Detection Framework for Pancreatic Cancer via CT Images
}

\author{Zhengdong Zhang$^{1}$, Shuai Li$^{1,2,*}$, Ziyang Wang$^{3}$ and Yun Lu$^{4}$% <-this % stops a space
\thanks{$^{*}$Shuai Li (e-mail: lishuai@buaa.edu.cn) is the corresponding author.}
\thanks{$^{1}$Zhengdong Zhang and Shuai Li are with State Key Laboratory of Virtual Reality Technology and Systems, Beihang University, Beijing 100191, China.}
\thanks{$^{2}$Shuai Li is with Beijing Advanced Innovation Center for Biomedical Engineering, Beihang University, Beijing 100191, China.}
\thanks{$^{3}$Ziyang Wang is with the Department of Computer Science, University of Oxford, Oxford OX1 3QD, the United Kingdom.}
\thanks{$^{4}$Yun Lu is with Shandong Key Laboratory of Digital Medicine and Computer Assisted Surgery, The Affiliated Hospital of Qingdao University, Qingdao 266003, China.}
}

\begin{document}

\maketitle
\thispagestyle{empty}
\pagestyle{empty}

%%%%%%%%%%%%%%%%%%%%%%%%%%%%%%%%%%%%%%%%%%%%%%%%%%%%%%%%%%%%%%%%%%%%
\begin{abstract}

%This electronic document is a ÒliveÓ template. The various components of your paper [title, text, heads, etc.] are already defined on the style sheet, as illustrated by the portions given in this document.
%\newline

%\indent \textit{Clinical relevance}— This is a brief additional statement on why a this might be of interest to practicing clinicians. Example: This establishes the anesthetic efficacy of 10\% intraosseous injections with epinephrine to positively influence cardiovascular function.
As Deep Convolutional Neural Networks (DCNNs) have shown robust performance and results in medical image analysis, a number of deep-learning-based tumor detection methods were developed in recent years. Nowadays, the automatic detection of pancreatic tumors using contrast-enhanced Computed Tomography (CT) is widely applied for the diagnosis and staging of pancreatic cancer. Traditional hand-crafted methods only extract low-level features. Normal convolutional neural networks, however, fail to make full use of effective context information, which causes inferior detection results. In this paper, a novel and efficient pancreatic tumor detection framework aiming at fully exploiting the context information at multiple scales is designed. More specifically, the contribution of the proposed method mainly consists of three components: Augmented Feature Pyramid networks, Self-adaptive Feature Fusion and a Dependencies Computation (DC) Module. A bottom-up path augmentation to fully extract and propagate low-level accurate localization information is established firstly. Then, the Self-adaptive Feature Fusion can encode much richer context information at multiple scales based on the proposed regions. Finally, the DC Module is specifically designed to capture the interaction information between proposals and surrounding tissues. Experimental results achieve competitive performance in detection with the AUC of 0.9455, which outperforms other state-of-the-art methods to our best of knowledge, demonstrating the proposed framework can detect the tumor of pancreatic cancer efficiently and accurately. 
\end{abstract}

%%%%%%%%%%%%%%%%%%%%%%%%%%%%%%%%%%%%%%%%%%%%%%%%%%%%%%%%%%%%%%%%%%%%%%%%%%%%%%%%
\section{INTRODUCTION}

%This template provides authors with most of the formatting specifications needed for preparing electronic versions of their papers. All standard paper components have been specified for three reasons: (1) ease of use when formatting individual papers, (2) automatic compliance to electronic requirements that facilitate the concurrent or later production of electronic products, and (3) conformity of style throughout a conference proceedings. Margins, column widths, line spacing, and type styles are built-in; examples of the type styles are provided throughout this document and are identified in italic type, within parentheses, following the example. Some components, such as multi-leveled equations, graphics, and tables are not prescribed, although the various table text styles are provided. The formatter will need to create these components, incorporating the applicable criteria that follow.

Since the Convolutional Neural Network was applied to visual data analysis~\cite{kohonen1982self}, there has been great progress in deep learning, computer vision and medical image processing. There is the potential to apply deep-learning-based approaches on medical image analysis, such as the tumor detection of pancreatic cancer. Pancreatic cancer is one of a malignant tumor diseases with around 7\% in 5-year survival rates~\cite{ryan2014pancreatic}~\cite{bray2018global}. The pancreas is a small organ located in the deep of human body, so that the difficulty of detection is significantly increased. Furthermore, missing the optimal time for radical surgery is the major cause of cancer death. CT imaging, a medical monitoring technology, which collects information of the tumor location, size and morphology, is helpful for the diagnosis and staging of pancreatic cancer compared with ultrasound imaging and Magnetic Resonance Imaging (MRI)~\cite{chu2019application}. Nevertheless, manually diagnosing requires doctors with rich clinical experience, because the quality of CT images varies between different CT scanners or operators, and pathological texture features are hard to be distinguished. Therefore, there is a growing need of studying on proposing a robust deep-learning-based algorithm for accurate pancreatic tumor detection.

Kishor achieved detection of pancreatic cancer in 2015~\cite{reddy2015detection}. A K-means clustering approach was utilized firstly to group the region of interests (ROIs). Then, A Haar wavelet transformation and threshold were adopted to classify images. His algorithm could be briefly deployed for the computer-aided system, whereas the performance of segmentation and classification might be seriously influenced by cancer pathological features. Li utilized saliency maps and densely-connected convolutional networks for pancreatic ductal adenocarcinoma diagnosis in 2019~\cite{li2019differential}. The high-level features were extracted and mapped to different types of pancreatic cysts. A larger training dataset might improve the performance. An approach for pancreatic tumor characterization inspired by the radiologists' inspiration and label proportion was illustrated by Sarfaraz~\cite{hussein2019lung}. He designed a 3D CNN-based graph-regularized sparse multi-task framework with a proportion-SVM to avoid the limited labeled data. It achieved sensitivity in diagnosing Intraductal Papillary Mucinous Neoplasms, but deep learning approaches such as Generative Adversarial Networks may show better performance~\cite{gutmann2012noise}.

Following the above consideration, an advanced framework for detecting human pancreatic tumor via CT images is proposed. Feature Pyramid Networks (FPN) utilize a top-down path with lateral connections to propagate semantic features in low levels, whereas the propagation through a long way increases the difficulty of exploiting accurate localization information~\cite{lin2017feature}. Therefore, a bottom-up Augmented Feature Pyramid aiming at shortening the information path and propagating low-level features is created at first. Secondly, Self-adaptive Feature Fusion to adaptively encode and integrate context information at multiple scales based on the proposals is designed, because the size of tumor is relatively small and nonuniform. Thirdly, inspired by the Non-local Neural Networks~\cite{wang2018non}, we employ a Dependencies Computation Module to compute dependencies and acquire interaction information with surrounding tissues. The expressiveness of features is enhanced by calculating the dependencies ranging from local to global. Subsequently, evaluation is illustrated and applied. The results achieve competitive performance compared with other deep-learning-based approaches.

\begin{figure*}[t]
	\centering
	\includegraphics[scale=0.55]{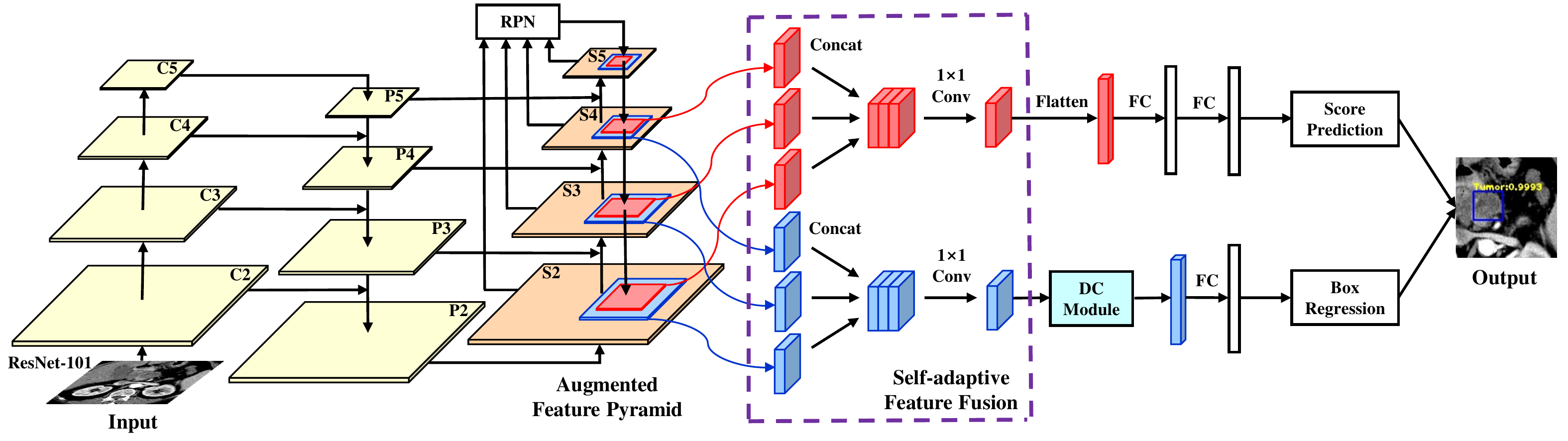}
	\caption{The architecture of the pancreatic tumor detection network.}
	\label{fig:framework}
\end{figure*}
\section{Methods}

The novel and efficient tumor detection framework we proposed is illustrated in Fig.~\ref{fig:framework}. The network utilizes FPN combined with Faster R-CNN~\cite{ren2015faster} as the backbone and the contribution of the proposed method consists of three components: Augmented Feature Pyramid networks, Self-adaptive Feature Fusion and a Dependencies Computation Module. Firstly, we feed the preprocessed CT images into the pre-trained ResNet-101 for feature extraction~\cite{he2016deep}, and then build the feature pyramid via up-sampling and lateral connections. Secondly, in order to enhance the entire feature hierarchy for improving detection performance, a bottom-up path is established to make low-level localization information propagation more efficient. Thirdly, we employ a Region Proposal Network (RPN) on each level to generate proposals~\cite{ren2015faster}, and then use Self-adaptive Feature Fusion to enlarge the corresponding ROIs and encode richer context information at multiple scales. Besides, we conduct the Dependencies Computation Module to capture dependencies with surrounding tissues of each proposal. Finally, detection results are predicted via a Score Prediction layer and a Box Regression layer, respectively. 

\subsection{Augmented Feature Pyramid Networks}

%First, confirm that you have the correct template for your paper size. This template has been tailored for output on the US-letter paper size. 
%It may be used for A4 paper size if the paper size setting is suitably modified.
In the process of feature extraction, DCNNs can extract semantic information. Meanwhile, high-level feature maps strongly respond to global features, which are beneficial to detect large objects~\cite{liu2018path}. The tumor, however, is relatively small in CT images, thus the consecutive pooling layers may lose the important spatial details of feature maps. In addition, the low-level accurate localization information is essential for tumor detection, but the information propagation path in FPN, which consists of more than 100 layers, affects the transmission effect. To this end, we build a bottom-up Augmented Feature Pyramid. As shown in Fig.~\ref{fig:framework}, firstly, we generate $\left\{{P2, P3, P4, P5}\right\}$ based on FPN. Then, the augmented path is established from the level $P2$, and $P2$ is directly used as $S2$, without any processing. Next, a $3\times3$ convolutional operator with stride 2 is conducted on a higher resolution feature map $S_i$ to reduce the map size. The down-sampled feature map is then merged with a coarser feature map $P_{i+1}$ by element-wise sum. In addition, we employ another $3\times3$ convolutional operator on each fused feature map to generate $S_{i+1}$ for following feature map generation. This process is iterated until the level $P5$ is used. In this way, we can acquire a new Augmented Feature Pyramid consisting of $\left\{{S2, S3, S4, S5}\right\}$.

\subsection{Self-adaptive Feature Fusion}

%The template is used to format your paper and style the text. All margins, column widths, line spaces, and text fonts are prescribed; please do not alter them. You may note peculiarities. For example, the head margin in this template measures proportionately more than is customary. This measurement and others are deliberate, using specifications that anticipate your paper as one part of the entire proceedings, and not as an independent document. Please do not revise any of the current designations
After acquiring the proposed regions by RPN, ROIs are assigned to one certain level according to their size, and the subsequent operations are performed on the same level, resulting in some useful information from other levels are discarded. In this case, instead of using a regression function to make predictions directly on the assigned proposals, we design a Self-adaptive Feature Fusion module, which aggregates hierarchical feature maps from multiple levels, to make full use of context information at multiple scales. Formally, the ROI with width $w$ and height $h$ is assigned to the level $S_k$ of the Augmented Feature Pyramid for each proposal by:
\begin{equation}
k=min(S_{max},max(\lfloor {k_0+\log_2{(\sqrt{wh}/C)}} \rfloor,S_{min}))\label{level}
\end{equation} 
In Eq.~\ref{level}, $C$ is the ImageNet pre-training size 224~\cite{krizhevsky2012imagenet}, and $k_0$ is set to 5, representing the coarsest level $S_5$. $S_{max}$ is 4, representing the level $S_4$. $S_{min}$ is 3, representing the level $S_3$. 

\begin{figure}[b]
	\centering
	\includegraphics[scale=0.732]{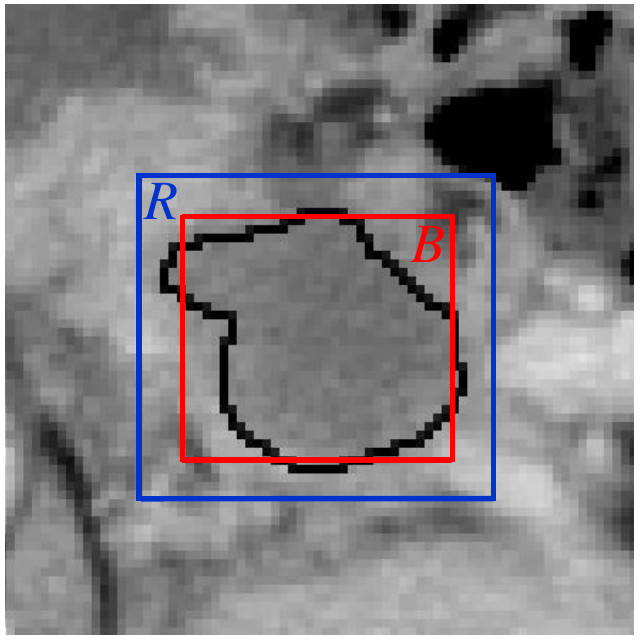}
	\caption{The example CT image of the given ROI $B$ and its corresponding region $R$.}
	\label{fig:enlarge}
\end{figure}

As shown in Fig.~\ref{fig:enlarge}, given an input ROI $B$, the predicted bounding box in red fails to cover the entire area of the tumor, especially the edge response, which results in information loss. To tackle this problem, we enlarge the width and height of ROI by the factor $S_w=$ 1.2 and $S_h=$ 1.2 to create a new region $R$ in blue. The new region $R$ contains richer context information, especially the responses about edges, which are strong indicators to accurate localization. Furthermore, as high-level features have larger receptive fields and capture more semantic information, low-level features have higher resolution and contain accurate localization details, which are complementary to abstract features. Both of them can help improve the detection performance, therefore, the regions $B$ and $R$ are mapped to the level $S_{k-1}$ and $S_{k+1}$, so that region $B$ and $R$ get three feature maps from three different scales, respectively. We employ 14$\times$14 ROI pooling over these maps to uniform the size. These descriptors are concatenated together and dimensions are reduced by 1$\times$1 convolutional operators. Finally, the $B$ based descriptor is used for score prediction, and the $R$ based descriptor is used for bounding box regression.
\subsection{Dependencies Computation Module}

In clinical practice, doctors pinpoint tumors through CT images by analyzing the global context information, local geometry structures, shape variations, and especially the spatial relations with surrounding tissues. In this case, we employ the Dependencies Computation Module to compute the response at a position, which is a weighted sum of the features at all positions on the enlarged region $R$. This operation can enable the network to pay more attention to the interactions and dependencies ranging from local to global, which is one of the most useful information for tumor detection. Specifically, given an input $x$, the entire Dependencies Computation Module is defined as follows:
\begin{equation}
y_i=softmax(\phi(x_i,x_j))h(x_j)\label{test}
\end{equation}
\begin{equation}
softmax(\phi(x_i,x_j))=\frac{exp(\phi(x_i,x_j))}{\sum_{j=1}^Nexp(\phi(x_i,x_j))}\label{softmax}
\end{equation}
In Eq.~\ref{test} and Eq.~\ref{softmax}, $i$ is the index of the chosen position, $j$ is the index of all other positions. The dependencies between any two positions are calculated via $\phi(x_i,x_j)=x_i^{T}W_{f}^{T}W_{g}x_j$, and $h(x_j)=W_hx_j$. $W_f$, $W_g$ and $W_h$ are matrices implemented by 1$\times$1 convolutional operators to reduce the number of channels. As the shape of the input feature $R$ is 14$\times$14$\times$512, the shape of three corresponding outputs is 14$\times$14$\times$256. At last, we use an addition operator to fuse it with the original feature, which denotes:
\begin{equation}
z_{i}=W_{z}y_{i}+x_i\label{eq}
\end{equation}
where $y_{i}$ is calculated in Eq.~\ref{test}, $x_i$ is the original input. $W_z$ is a 1$\times$1 convolution layer used to restore the shape back to 14$\times$14$\times$512.

\begin{figure}[t]
	\centering
	\includegraphics[scale=0.542]{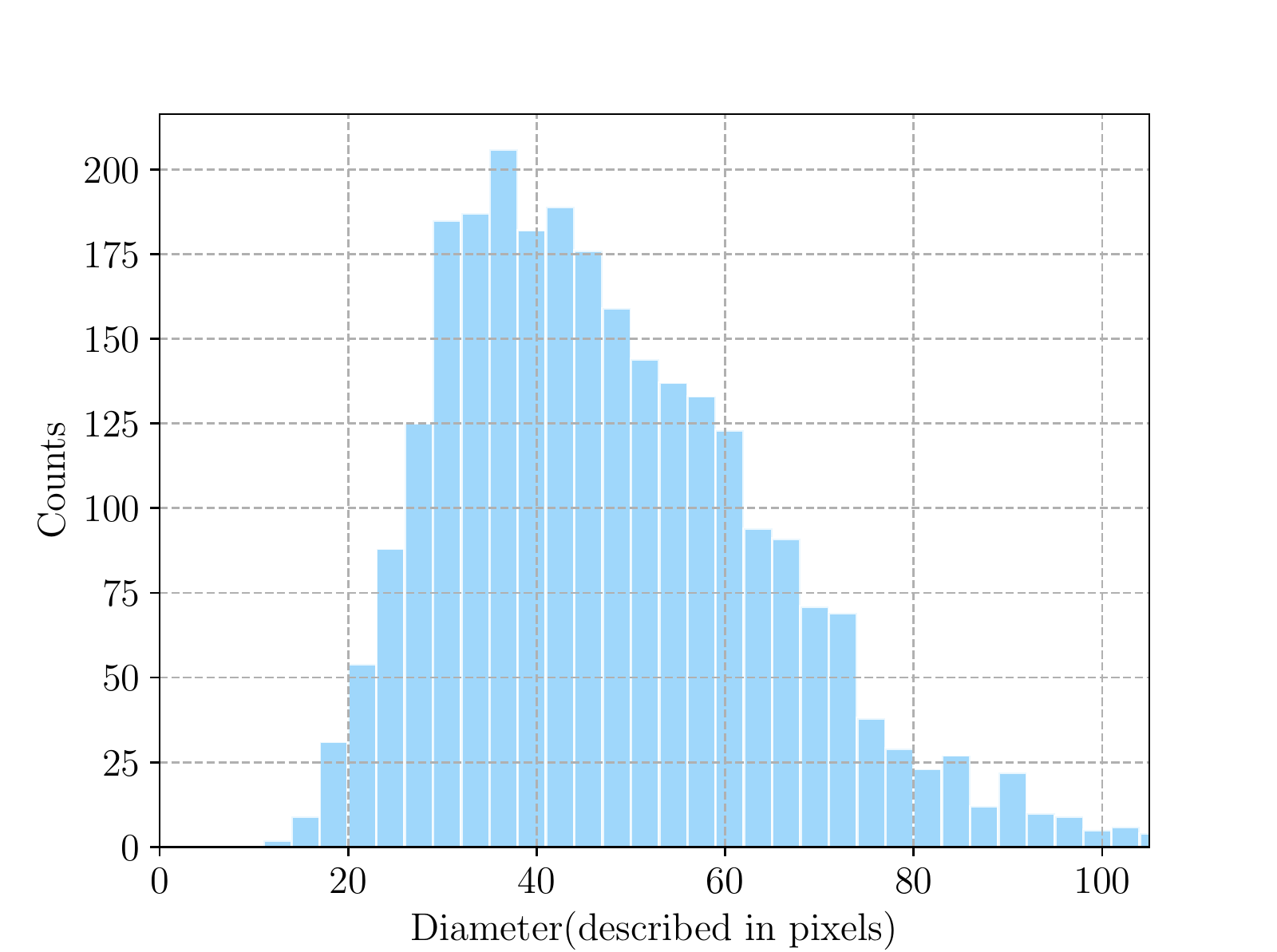}
	\caption{The Histogram of the diameter of tumor in the dataset.}
	\label{fig:diameter}
\end{figure}

\section{EXPERIMENTS AND RESULTS}

%Before you begin to format your paper, first write and save the content as a separate text file. Keep your text and graphic files separate until after the text has been formatted and styled. Do not use hard tabs, and limit use of hard returns to only one return at the end of a paragraph. Do not add any kind of pagination anywhere in the paper. Do not number text heads-the template will do that for you.

%Finally, complete content and organizational editing before formatting. Please take note of the following items when proofreading spelling and grammar:

\subsection{Dataset} 
%Define abbreviations and acronyms the first time they are used in the text, even after they have been defined in the abstract. Abbreviations such as IEEE, SI, MKS, CGS, sc, dc, and rms do not have to be defined. Do not use abbreviations in the title or heads unless they are unavoidable.
The model is trained by a dataset of pancreatic CT images provided by The Affiliated Hospital of Qingdao University. The dataset contains 2890 CT images, in which 2650 images are for training, and 240 images are for testing. There is no overlap between the training set and the test set, and all the images are labeled by three experienced doctors with accurate bounding boxes. The diameter distribution of the tumor in the dataset is illustrated in Fig.~\ref{fig:diameter}. The diameter ranges from 15 to 104 pixels, and most of them are between 20 and 80 pixels. We preprocess these images and conduct data augmentation, including horizontal flip, vertical flip and diagonal flip before training.  
\subsection{Experiment Setup}

The proposed method is implemented in Python using Tensorflow. During the training process, we set the batch size to 1, the momentum is 0.9 and weight decay is 0.0001, the learning rate is 0.001 for the first 30K iterations, 0.0001 for the next 20K and 0.00001 for the last 10K. For each mini-batch, we sample 512 ROIs with positive-to-negative ratio 1:1. For anchors, according to the tumors' diameter distribution illustrated in Fig.~\ref{fig:diameter}, we choose 5 scales with box areas of $16^2$, $32^2$, $64^2$, $128^2$, $256^2$, and 5 anchor ratios of 1:1, 1:1.5, 1.5:1, 1:2 and 2:1. The hardware settings are Intel (R) Core i7-9800X CPU, Nvidia GeForce RTX2080 Ti GPU and 32GB memory on Ubuntu 64bits Linux desktop.  
%\begin{itemize}

%\item Use either SI (MKS) or CGS as primary units. (SI units are encouraged.) English units may be used as secondary units (in parentheses). An exception would be the use of English units as identifiers in trade, such as Ò3.5-inch disk driveÓ.
%\item Avoid combining SI and CGS units, such as current in amperes and magnetic field in oersteds. This often leads to confusion because equations do not balance dimensionally. If you must use mixed units, clearly state the units for each quantity that you use in an equation.
%\item Do not mix complete spellings and abbreviations of units: ÒWb/m2Ó or Òwebers per square meterÓ, not Òwebers/m2Ó.  Spell out units when they appear in text: Ò. . . a few henriesÓ, not Ò. . . a few HÓ.
%\item Use a zero before decimal points: Ò0.25Ó, not Ò.25Ó. Use Òcm3Ó, not ÒccÓ. (bullet list)

%\end{itemize}
\begin{figure}[b]
	\centering
	\includegraphics[scale=0.59]{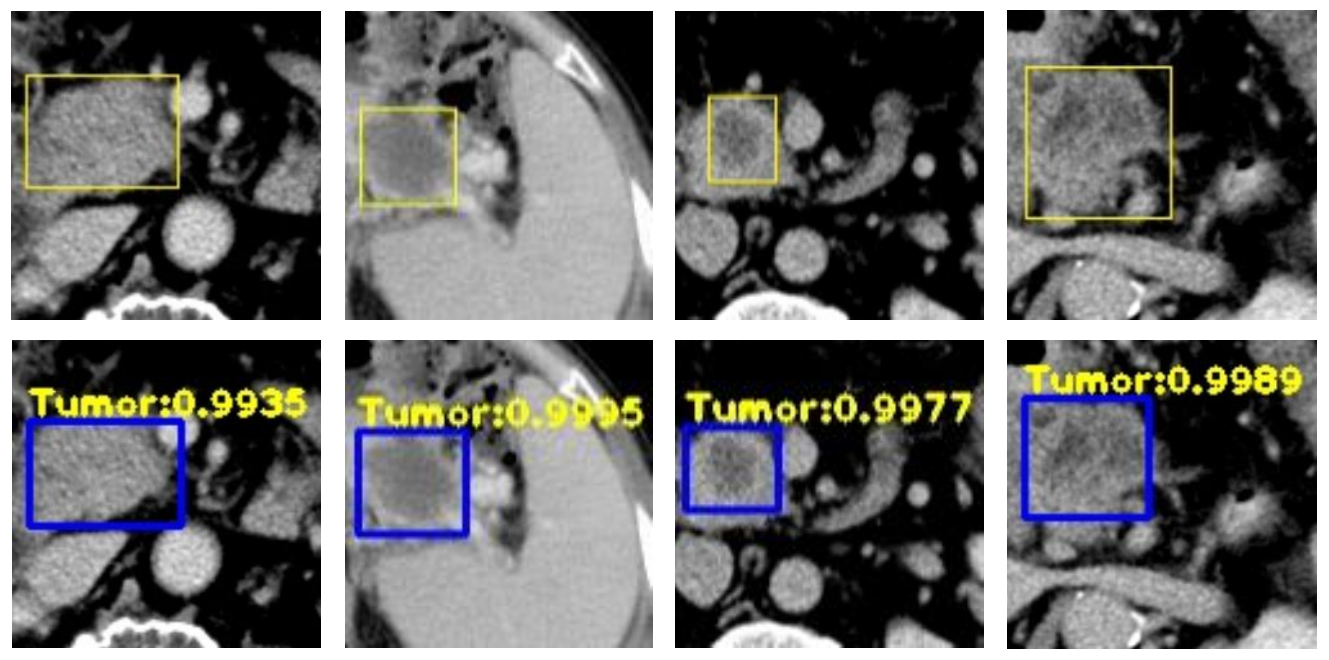}
	\caption{Example results of tumor detection. The first row are the ground truth, the second row are the corresponding detection results of the proposed method.}
	\label{fig:result}
\end{figure}

\subsection{Results and Discussion}

\begin{figure}[h]
	\centering
	\includegraphics[scale=0.5293]{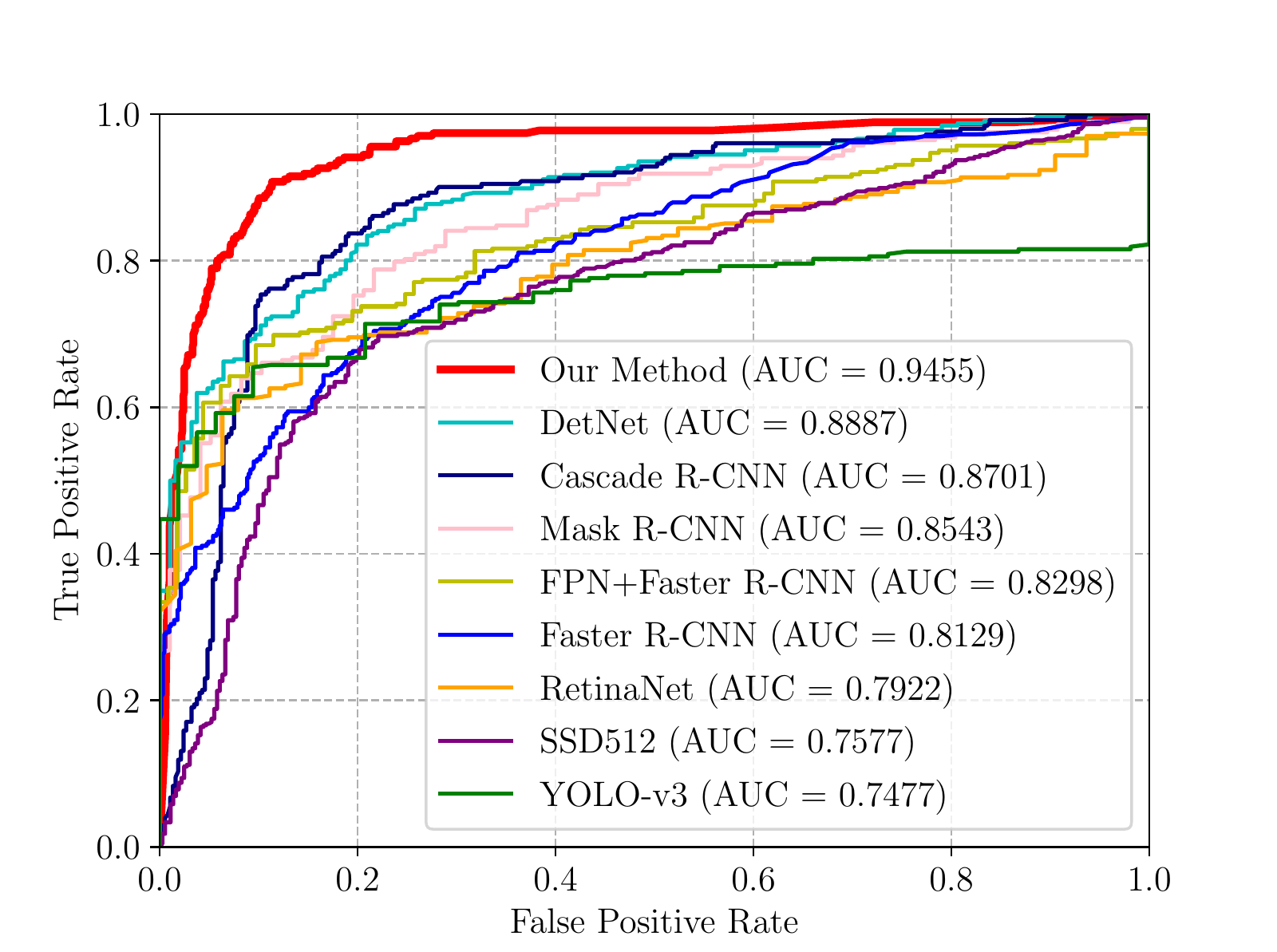}
	\caption{The ROC curves of different methods for pancreatic tumor detection.}
	\label{fig:roc}
\end{figure}

Example results of tumor detection are shown in Fig.~\ref{fig:result}, the localization is relatively accurate and the corresponding probability score is high as well. In order to evaluate the detection performance, the proposed method is compared with classical object detection algorithms, including DetNet~\cite{Li2018DetNet}, Cascade R-CNN~\cite{cai2018cascade}, Mask R-CNN~\cite{he2017mask}, FPN~\cite{lin2017feature}, Faster R-CNN~\cite{ren2015faster}, RetinaNet~\cite{lin2017focal}, SSD512~\cite{liu2016ssd} and YOLO-v3~\cite{redmon2018yolov3}. These algorithms are trained and tested using the same pancreatic CT dataset without additional modifications. The Intersection Over Union (IOU) between predicted bounding box $B_p$ and the corresponding ground-truth bounding box $B_{gt}$ is calculated for each result, which is defined as follows:
\begin{equation}
IOU=\frac{B_{gt}\cap{B_p}}{B_{gt}\cup{B_p}}\label{eq}
\end{equation}
Furthermore, the detection results whose IOU are higher than 0.5 are regarded as valid results. As shown in Table~\ref{table:performance}, the proposed method achieves the best 0.8376, 0.9179 and 0.9018 in Sensitivity, Specificity and Accuracy, respectively, outperforming other methods by a notable margin. The corresponding Receiver Operating Characteristics (ROC) curves in Fig.~\ref{fig:roc} show that our proposed method is superior to other methods with the Area Under Curve (AUC) of 0.9455. 
\begin{table}[t]
	\caption{Detection performance comparison among different algorithms on test set}
	\def\arraystretch{1.21}
	\begin{center}
		\begin{tabular}{|p{3.0cm}<{\centering}|p{1.2cm}<{\centering}|p{1.2cm}<{\centering}|p{1.2cm}<{\centering}|}
			%\hline
			%\textbf{Table}&\multicolumn{3}{|c|}{\textbf{Table Column Head}} \\
			%\cline{2-4} 
			\hline
			\textbf{Methods} & \textbf{Sensitivity}& \textbf{Specificity}& \textbf{Accuracy} \\
			\hline
			SSD512~\cite{liu2016ssd}& 0.4238& 0.9088& 0.6411  \\
			FPN + Faster R-CNN~\cite{lin2017feature}& 0.6984& 0.8584& 0.7416 \\
			YOLO-v3~\cite{redmon2018yolov3}& 0.7697& 0.5849& 0.7423  \\
			Mask R-CNN~\cite{he2017mask}& 0.7244& 0.8247& 0.7500  \\
			Faster R-CNN~\cite{ren2015faster}& 0.4877& 0.9131& 0.7538  \\
			DetNet~\cite{Li2018DetNet}& 0.6932& 0.9032& 0.7695 \\
			RetinaNet~\cite{lin2017focal}& 0.8245& 0.5238& 0.7726 \\
			Cascade R-CNN~\cite{cai2018cascade}& 0.6309& 0.9113& 0.7981  \\
			\hline
			\textbf{Our Method} & \textbf{0.8376}& \textbf{0.9179}& \textbf{0.9018} \\
			\hline
		\end{tabular}
		\label{table:performance}
	\end{center}
\end{table}

In addition, in order to evaluate the proposed method more accurately, Free-Response Receiver Operating Characteristics (FROC) is used to compute the Sensitivity at 7 FP/scan rates. Our proposed method achieves an average score of 0.901, the corresponding results are documented in Table~\ref{table:froc}.

\begin{table}[t]
	\tiny
	\caption{Detection performance in terms of Sensitivity based on different FPs/scan rates on test set}
	\def\arraystretch{1.8}
	\begin{center}
		%\begin{tabular}{|p{1.0cm}<{\centering}|p{0.4cm}<{\centering}|p{0.5cm}<{\centering}|p{0.5cm}<{\centering}|p{0.5cm}<{\centering}|p{0.5cm}<{\centering}|p{0.5cm}<{\centering}|p{0.5cm}<{\centering}|p{0.5cm}<{\centering}|p{0.5cm}<{\centering}|}
		\begin{tabular}{|c|c|c|c|c|c|c|c|c|}
			%\hline
			%\textbf{Table}&\multicolumn{3}{|c|}{\textbf{Table Column Head}} \\
			%\cline{2-4} 
			\hline
			\textbf{FPs/scan} & \textbf{0.125}& \textbf{0.25}& \textbf{0.5}& \textbf{1}& \textbf{2}& \textbf{4}& \textbf{8}& \textbf{Average}\\
			\hline
			\textbf{Sensitivity} & 0.671& 0.804& 0.907& 0.963& 0.977& 0.986& 0.998& \textbf{0.901}\\
			\hline
		\end{tabular}
		\label{table:froc}
	\end{center}
\end{table}
\begin{table}[h]
	\caption{Ablation studies on Effects comparison of the proposed components and their
		combinations}
	\def\arraystretch{1.24}
	\begin{center}
		\begin{tabular}{|c|c|c|c|}
			%\hline
			%\textbf{Table}&\multicolumn{3}{|c|}{\textbf{Table Column Head}} \\
			%\cline{2-4} 
			\hline
			\textbf{Augmented}&\textbf{Self-adaptive}& &  \\
			\textbf{Feature}&\textbf{Feature}&\textbf{DC Module}&\textbf{Accuracy} \\
			\textbf{Pyramid}&\textbf{Fusion}& &  \\
			\hline
			&  &  & \textbf{0.7416}   \\
			$\checkmark$&  &  & 0.8132 \\
			& $\checkmark$& & 0.8359 \\
			&  $\checkmark$& $\checkmark$& 0.8697 \\
			$\checkmark$& $\checkmark$&  & 0.8541 \\
			$\checkmark$& $\checkmark$& $\checkmark$ & \textbf{0.9018} \\
			\hline
		\end{tabular}
		\label{table:contribution}
	\end{center}
\end{table}
Extensive ablation experiments are conducted to analyze the effects of the proposed components and their combinations in our method. The results are documented in Table~\ref{table:contribution}, the Augmented Feature Pyramid networks and Self-adaptive Feature Fusion can significantly improve the accuracy in individual cases. Finally, the detection accuracy can be significantly improved from 0.7416 to 0.9018 by the combinations of these three proposed components.

\section{CONCLUSION}

In this paper, we study on how to accurately detect the tumor of pancreatic cancer, which is of great significance for the diagnosis in clinical practice. We establish an Augmented Feature Pyramid to propagate low-level accurate localization information. We also design Self-adaptive Feature Fusion to capture richer context information at multiple scales. Finally, we compute the relation information of the features via the Dependencies Computation Module. Comprehensive evaluations and comparisons are completed, and our proposed method achieves promising performance. In the future, we will continue studying on the staging of pancreatic cancer to assist the doctor's clinical diagnosis. %A conclusion section is not required. Although a conclusion may review the main points of the paper, do not replicate the abstract as the conclusion. A conclusion might elaborate on the importance of the work or suggest applications and extensions. 

\addtolength{\textheight}{-12cm}   % This command serves to balance the column lengths

%%%%%%%%%%%%%%%%%%%%%%%%%%%%%%%%%%%%%%%%%%%%%%%%%%%%%%%%%%%%%%%%%%%%%%%%%%%%%%%%

%%%%%%%%%%%%%%%%%%%%%%%%%%%%%%%%%%%%%%%%%%%%%%%%%%%%%%%%%%%%%%%%%%%%%%%%%%%%%%%%

%%%%%%%%%%%%%%%%%%%%%%%%%%%%%%%%%%%%%%%%%%%%%%%%%%%%%%%%%%%%%%%%%%%%%%%%%%%%%%%%
%\section*{APPENDIX}


\section*{ACKNOWLEDGMENT}

%The preferred spelling of the word ÒacknowledgmentÓ in America is without an ÒeÓ after the ÒgÓ. Avoid the stilted expression, ÒOne of us (R. B. G.) thanks . . .Ó  Instead, try ÒR. B. G. thanksÓ. Put sponsor acknowledgments in the unnumbered footnote on the first page.

%%%%%%%%%%%%%%%%%%%%%%%%%%%%%%%%%%%%%%%%%%%%%%%%%%%%%%%%%%%%%%%%%%%%%%%%%%%%%%%%

%References are important to the reader; therefore, each citation must be complete and correct. If at all possible, references should be commonly available publications.

This research is supported in part by Foundation of Shandong provincial Key Laboratory of Digital Medicine and Computer assisted Surgery (SDKL-DMCAS-2018-01), National Natural Science Foundation of China (NO. 61672077 and 61532002), Applied Basic Research Program of Qingdao (NO. 161013xx), and Beijing Natural Science Foundation-Haidian Primitive Innovation Joint Fund (L182016).

\bibliographystyle{ieeeconf}
\bibliography{ref}

\end{document}